\documentclass{article}

\usepackage{PRIMEarxiv}

\usepackage[utf8]{inputenc} % allow utf-8 input
\usepackage[T1]{fontenc}    % use 8-bit T1 fonts
\usepackage{hyperref}       % hyperlinks
\usepackage{url}            % simple URL typesetting
\usepackage{booktabs}       % professional-quality tables
\usepackage{amsfonts}       % blackboard math symbols
\usepackage{nicefrac}       % compact symbols for 1/2, etc.
\usepackage{microtype}      % microtypography
\usepackage{lipsum}
\usepackage{fancyhdr}       % header
\usepackage{graphicx}       % graphics
\graphicspath{{media/}}     % organize your images and other figures under media/ folder

\title{Depth-guided Free-space Segmentation for a Mobile Robot
}

\author{
    Christos Sevastopoulos$^{1,2}$, Joey Hussain$^1$, Stasinos Konstantopoulos$^2$, Vangelis Karkaletsis$^2$, Fillia Makedon$^1$ \\
    $^1$Department of Computer Science \& Computer Engineering, University of Texas at Arlington, Arlington, TX, USA \\
    $^2$Institute of Informatics and Telecommunications, NCSR 'Demokritos', Agia Paraskevi, Attica, Greece \\
    \{christos.sevastopoulos,joey.hussain\}@mavs.uta.edu \\
    \{konstant,vangelis\}@iit.demokritos.gr \\
    makedon@uta.edu
}

\pdfoutput=1

\begin{document}
\maketitle

\begin{abstract}
Accurate indoor free-space segmentation is a challenging task due to the complexity and the dynamic nature that indoor environments exhibit. We propose an indoors free-space segmentation method that  associates large depth values with navigable regions. Our method leverages an unsupervised masking technique that, using positive instances, generates segmentation labels based on textural homogeneity and depth uniformity. Moreover, we generate superpixels corresponding to areas of higher depth and align them with features extracted from a Dense Prediction Transformer (DPT). Using the estimated free-space masks and the DPT feature representation, a SegFormer model is fine-tuned on our custom-collected indoor dataset.
Our experiments demonstrate sufficient performance in intricate scenarios characterized by cluttered obstacles and challenging identification of free space.
\end{abstract}

% keywords can be removed

\section{Introduction}

Robust free-space segmentation is a fundamental task for mobile robots operating indoors. By correctly identifying areas devoid of obstacles, robots can navigate safely and efficiently, avoiding collisions~\cite{b28}. However, achieving accurate free-space segmentation remains challenging due to the complex and dynamic nature of indoor environments.

Contemporary research leverages deep learning techniques, with substantial potential for this task. By employing pre-trained models and transfer learning, mobile agents learn to identify traversable regions using positive instances representing obstacle-free areas~\cite{hirose2019deep},~\cite{b29}. Nevertheless, challenges such as the stochastic nature of the environment, presence of humans, variations in lighting conditions and occlusions can largely affect the perception of free space~\cite{b34}. Additionally, indoor scenes with ambiguous and semantically diverse objects, like furniture, household items, and interior structures may exhibit textures similar to traversable surfaces, and thus lead to inconsistent predictions.

In this research, our objective is to overcome these limitations by proposing a novel learning-based approach for indoor free-space segmentation. Our method goes beyond conventional object-centric models~\cite{b31,b30,b32, b33} by leveraging depth information to focus on precise detection and segmentation of free space. By utilizing the power of pre-trained transformer-based networks, transfer learning, and positive instances, our aim is to train a model that can identify safe and navigable areas within complex indoor environments. In order to achieve this, we propose that areas classified as traversable free-space are likely to exhibit greater depth values.

As an overview, the contributions of this work are the following:
\begin{itemize}
    \item We hypothesize that areas classified as traversable free-space exhibit greater depth values, forming a key criterion for accurate free-space segmentation
    \item An automated masking technique that leverages textural homogeneity and depth uniformity to produce meaningful segments
    \item Learning to predict indoor free-space segmentation using only positive instances for training
\end{itemize}

\begin{figure*}
    \centering
    \includegraphics[width=17cm]{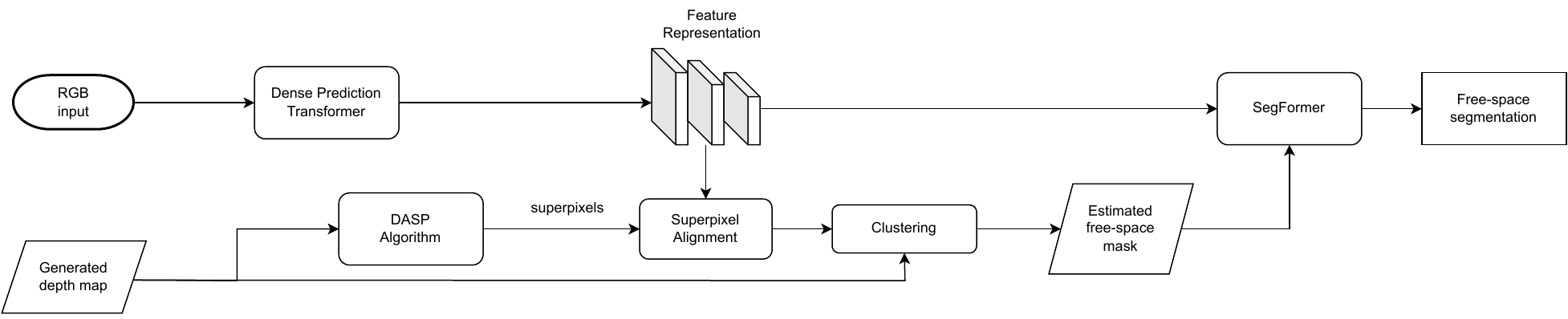}
    \caption{Proposed methodology }
    \label{fig:Arch}
\end{figure*}

\section{Related Work}

Extensive research has been conducted on predicting a scene's geometry using RGB-D information, primarily focusing on outdoor scenarios~\cite{b22,b25}, providing valuable insights into identifying the location of traversable free-space~\cite{wei2021perceive},~\cite{b24}. In indoor settings, notable attempts have been made to determine the scene's traversability only at a higher conceptual level~\cite{b21},~\cite{b26} without specifically delving into the segmentation of free-space.

In general, the task of free-space detection encompasses the utilization of diverse sensors like cameras, laser scanners, or occupancy grid maps, in conjunction with deep learning methodologies, to recognize and identify safe navigation areas within the environment~\cite{b27}. Pizzati and Garcia~\cite{b15} 
propose the use of a \emph{Convolutional Neural Network (CNN)}
architecture that can identify and classify free-space areas within different lanes of the road simultaneously.

The integration of RGB and depth information has demonstrated its advantages in the field of semantic segmentation. Cao et al.~\cite{b6} exploited estimated depth features to develop an RGB-D semantic segmentation approach that incorporates a multi-task training strategy.
Nekrasov et al.~\cite{b7} propose a method that adapts a real-time semantic segmentation network to handle multiple tasks with asymmetric datasets. Mousavian et al.~\cite{b8} address simultaneous depth estimation and semantic segmentation using deep CNNs by proposing a model trained independently for each task, followed by joint fine-tuning using a single loss function. 
A novel depth-aware convolution and depth-aware average pooling operations that utilize depth similarity between pixels, to incorporate geometric information into a CNN for RGB-D semantic segmentation is illustrated in the work of Wang and Neumann~\cite{b9}.

With the emergence of transformers, the
\emph{Vision Transformer (ViT)} architecture~\cite{b5} has been associated with a multitude of computer vision tasks due to its excellence at capturing global context and long-range dependencies within the input image. Strudel et al.~\cite{b10} extend the use of ViT, and introduce \emph{Segmenter}, a transformer-based model for semantic segmentation. Generally speaking, the hierarchical structure is highly suitable for tasks involving pixel-level predictions, such as semantic segmentation and object detection. SegFormer~\cite{b11} combines ViT-architecture with lightweight \emph{Multilayer Perceptron (MLP)} decoders, and has shown significant efficiency for semantic segmentation tasks. 
Other transformer-based models highlighting the diverse applicability of such in computer vision tasks can be found in~\cite{b12} with particular focus on image restoration, in~\cite{b13} addressing image classification, object detection, and semantic segmentation, and in~\cite{b14} specifically targeting the segmentation of medical imaging.

\section{Methodology}

\subsection{Overall approach}

Figure~\ref{fig:Arch} illustrates the proposed methodology's architecture.
Our approach is based on the central hypothesis that areas classified as traversable free-space will generally exhibit greater depth values. To validate this hypothesis, we focus on training a refined semantic segmentation model using positive RGB indoor instances that the robot can safely traverse, such as hallways and wide paths. Our method relies solely on the RGB data we collected, while for generating the depth maps, we utilized the depth estimation model provided by Niantic Labs~\cite{b16}, allowing to extract valuable depth information from the aforementioned RGB instances. 

Additionally, we propose an unsupervised mask generation framework that leverages depth information to capture textural homogeneity and depth uniformity as indicators of free space. Subsequently, we evaluate the performance of the model on challenging scenarios where free space is present yet restricted due to the presence of cluttered objects.

\subsection{Mask Annotation}

To generate annotation masks for free-space segmentation, we build upon the following main hypothesis: Free space is expected to be depicted by textural homogeneity in conjunction with depth uniformity. This means, that in order to search for free space in the image, we consider these particular superpixels that are characterized by similarity in both color and depth information. Therefore, we use a superpixel oversegmentation algorithm to generate unsupervised free-space masks by capturing intra-region similarities towards producing meaningful segments: the Depth-Adaptive Superpixels (DASP) algorithm~\cite{b1}.

Overall, our method is influenced by the findings and approaches outlined in~\cite{b2}, however we differentiate in the sense that "seed" superpixels are generated through the guidance of depth information. In lieu of empirically computing the parameters that define the location prior, we use the DASP algorithm to generate the seeds needed for segmentation.

\begin{figure}
    \centering
    \includegraphics[width=20cm]{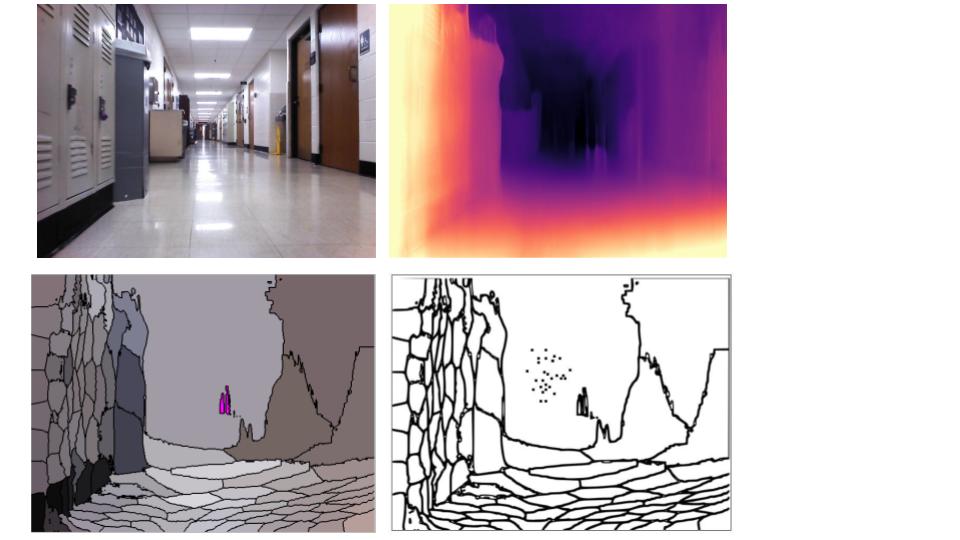}
    \caption{Upper row: RGB-D pair 
Bottom row: On the left, the DASP algorithm performs oversegmentation, dividing the image into superpixels. On the right, the seeds generated by the DASP algorithm, described by the dotted area, represent the area with the greatest depth}
    \label{fig:ovsp}
\end{figure}

\subsection{Features Extraction}

We intend to associate superpixel clusters with features that have been extracted by a Dense Prediction Transformer~\cite{b4}, that uses the \emph{Vision Transformer} (ViT) architecture as a backbone for dense prediction tasks such as semantic segmentation. DPT leverages the strengths of 1) Transformers, known for their ability to model long-range dependencies, and 2) Convolutional Neural Networks (CNNs) that can efficiently process and encode spatial information within the input images.
Consequently, DPT allows for efficient capture of both local and global information. We remove the last layer of the pre-trained Intel's DPT-Large\footnote{\url{https://github.com/isl-org/DPT}} model, and the resulting extracted features, in the form of a vector of size 577x1024, exhibit rich high-level semantic information as well as spatial details at a higher resolution.

\subsection{DASP}

DASP utilizes spectral graph theory to calculate segments through graph cuts. Essentially, it creates a graph where each pixel is represented as a node, and the edges represent the similarity between pixels. The similarity computation takes into account both color and depth information and this distinctive property is of fundamental importance for our method i.e. estimating segments that exhibit the largest depth.
An additional asset that DASP displays and can be substantial towards estimating the segments of higher importance, is the generation of segmentation seeds that are expected to be associated with free space. DASP achieves this by calculating the depth gradient of the input image and clustering it, to eventually obtain a collection of initial segmentation seeds. Hence, instead of relying on a location prior as in~\cite{b2}, we can utilize the seed coordinates that guide the segmentation towards areas of larger depth (Figure~\ref{fig:ovsp}).

\subsection{Superpixel alignment}

To establish a spatial association between the distinct features derived from the DPT and the coherent superpixels generated by the DASP algorithm, we use a sophisticated technique known as
\emph{superpixel alignment} inspired by the work by Tsutsui et al.~\cite{b2} and influenced by the concept of RoIAlign~\cite{b3}.

Firstly, a selection of ten representative locations is made within each superpixel, forming crucial anchor points within these coherent regions. Next, for each of these chosen anchor points, the algorithm identifies the four nearest neighbors of each selected pixel for the bilinear interpolation. This process creates a local context around each anchor point, facilitating a focused examination of the surrounding area. 

In order to effectively associate the features extracted by the DPT with the representative anchor points within superpixels, the technique of bilinear interpolation is implemented. Specifically, it estimates the features at the anchor points by considering the features of the nearby pixels, and thus seamlessly integrating the two sets of information. By projecting the features extracted from the neighboring pixels onto the representative anchor points using bilinear interpolation, a direct alignment between DPT features and superpixels is achieved. 

As a result, this alignment is crucial in ensuring that the information extracted by the DPT accurately corresponds to the meaningful regions represented by the superpixels. Afterwards, we employ average pooling to aggregate the features within each superpixel and therefore generate a  comprehensive representation of the superpixels' characteristics. 

% Overall, this superpixel alignment process ensures that the extracted features are effectively aligned with the corresponding regions of interest, enabling accurate analysis and segmentation of the free space areas.

\subsection{Superpixel clustering}
We aim to guide the segmentation towards traversable areas that exhibit the largest depth. Therefore, we cluster superpixels with respect to the semantic features and additionally, we select the cluster that corresponds to the area with the largest depth. As highlighted by Weikersdorfer et al.~\cite{b1}, the density of superpixel clusters in the image space is computed from the depth image. For detailed information on how this density is computed, the reader is referred to the original article.

The steps of the DASP algorithm can be summarized as follows:
1) the density of the depth-adaptive superpixel clusters is calculated based on the depth input image 2) an initial set of cluster centers is obtained by applying a Poisson disc sampling method 3) these sampled cluster centers are utilized in a density-adaptive local iterative clustering algorithm to assign pixels to their respective cluster centers. We anticipate that the superpixels representing open areas will have lower density values.This is because there are fewer superpixels/clusters required to represent those areas due to their relatively uniform or homogeneous nature. 

Consequently, our objective is to identify the clusters within the Poisson disk distribution that are most likely to define free space. For a given depth image, we estimate the superpixel density values, and using the k-means clustering technique, we single out the clusters. In a similar fashion to~\cite{b2}, the initial cluster is encouraged to include pixels associated with free space by positioning its cluster center as the average of features weighted by their spatial distribution. In contrast, the rest of the clusters have a repelling weight assigned to their members, prompting them to spread out spatially from their previous location.

\subsection{Fine-tune a SegFormer}

Using the masks produced at the previous step, we fine-tune a SegFormer model on our custom-collected dataset. The SegFormer is merging the Transformer's architectural backbone with a lightweight decoder. In contrast to ViT, its \emph{Mix Transformer} encoder (MiT) does not use any positional encodings and can generate multi-level feature maps (both high-resolution fine features and low-resolution coarse features) due to its hierarchical structure. Furthermore, a series of lightweight (fewer number of parameters) Multi-Layer Perceptrons (MLP) is used as a decoder, that exhibits the attribute of combining both local and global attention, and eventually creates powerful and meaningful representations accompanied by strong generalization capabilities. This last aspect, along the dominant multi-scale feature learning ability of the SegFormer's encoder, can be proven to be crucial for our method; specifically, we use a SegFormer-B0 pre-trained on ImageNet-1K, and replace the classification head for fine-tuning on our dataset.

\section{Experimental Setup}

\subsection{Data Collection}

For the data collection, a human operator directly teleoperated the Summit-XL Steel robotic platform with a wireless PS4 controller. We used ROS Melodic\footnote{\url{http://wiki.ros.org/melodic}}
and the \textit{message\_filters} data synchronization package\footnote{\url{http://wiki.ros.org/message filters}}.
to operate the robot and to record the
RGB, laser range finder, IMU, and encoder channels.

\begin{figure}
  \centering
  \begin{minipage}[b]{0.47\linewidth}
    \centering
    \includegraphics[width=\linewidth]{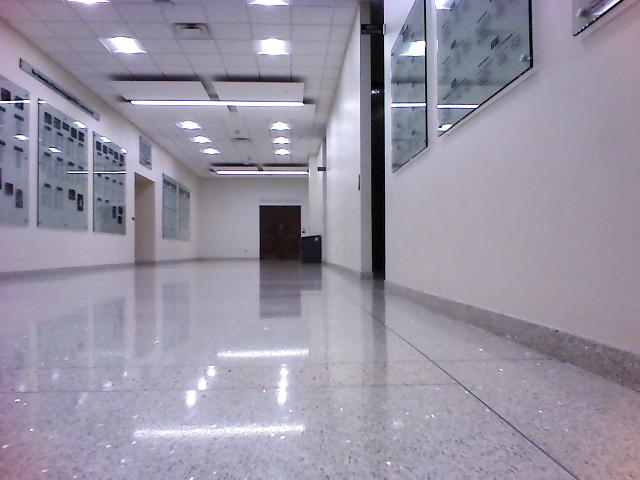}
    %\caption{}
    \label{fig:image1}
  \end{minipage}
  \hfill
  \begin{minipage}[b]{0.47\linewidth}
    \centering
    \includegraphics[width=\linewidth]{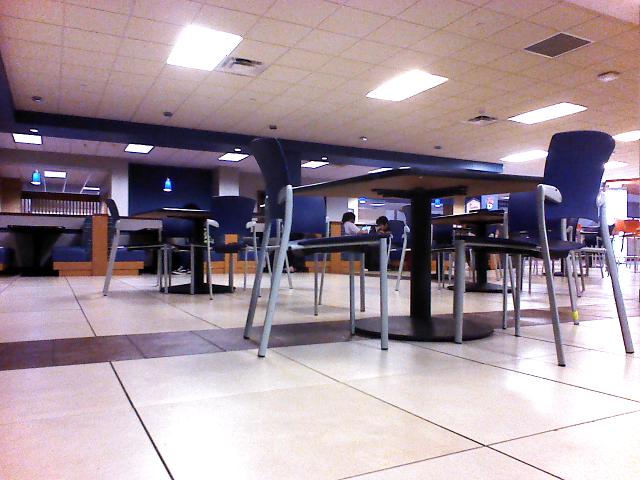}
    %\caption{}
    \label{fig:image2}
  \end{minipage}
  \caption{Illustrative examples of positive and challenging instances. For the positive example (left) the robot is free to traverse while in the challenging case (right), the free space is limited due to the presence of cluttered objects}
  \label{fig:examples}
\end{figure}

\subsection{Data Collection and Annotation}

The robot was deployed in various buildings on the University of Texas, Arlington (UTA) campus, where it navigated hallways and areas with changing lighting conditions and diverse objects. To ensure safety, the operator stopped the robot if there was a potential interference with humans or objects. In total, we collected a dataset of 3324 monocular RGB images.

Our primary objective was to utilize the Positive-Unlabeled (PU) method, described in \cite{b49}, to divide each source set (corresponding to a specific university building) into two subsets: positive and unlabeled image instances. However, relying solely on wheel velocity for automated annotation could lead to inaccuracies, especially in scenarios with unexpected human presence or limited odometry information. To enhance the reliability of annotation, we incorporated information from a laser scanner.

The annotation process involved examining the velocities of all four wheels, establishing a threshold value of 1 m/s, and a time window of 2.5 seconds. We defined three conditions: the wheel velocities consistently exceeding the threshold, the robot moving forwards or turning, and no obstacles detected within a 1.2-meter range of the laser sensor. If these conditions were met, we labeled the central frame within the time window as "positive" representing a traversable scene. Frames that did not meet the conditions were left unlabeled. From the unlabeled set, we manually removed scenes where the robot had limited or no free space to proceed, and labeled scenes with both free space and obstacles as "challenging". Thus, our dataset for experiments consisted of 2553 images, including 2010 positive and 443 challenging instances.

\section{Results and discussion}

\subsection{Implementation Details}

We used the PyTorch\footnote{\url{https://pytorch.org}} framework as the basis of our experiments. Training was done on a machine with 2 Titan RTX GPUs (24GB GDDR6 RAM, 4608 Cuda Cores). Using the standard cross-entropy loss function, we trained for 50 epochs unless an early stopping callback terminated the trial upon observed convergence. As training parameters we used: batch size = 16, learning rate = 0.01 and weight decay = 5e-4. The fine-tuning of the layers was accomplished using stochastic gradient descent (SGD). All images' initial dimensions of 640x480 pixels were re-scaled to 224x224 using the default PyTorch interpolation. In the following sections, we present the best results achieved when using 1800 of positive instances for training, and 453 challenging instances for testing.

\subsection{Performance Analysis}

We train our method on the positive instances and test on the challenging ones. For our experiments, the standard semantic segmentation metrics, Intersection over Union (IoU) is used and we found that the best results were achieved for k=5 clusters.

\begin{table}[h]
\caption{Performance of each method given different inputs}
\centering

\begin{tabular}{cccccc}

\hline
\textbf{Method} & \textbf{RGB} & \textbf{Depth} & \textbf{Training Data} & \textbf{Model} & \textbf{IoU Score} \\
\hline

\cite{b2}  & \checkmark & $\times$ &Automated &SegNet& 0.773   \\ 
\cite{b2}  & \checkmark  &  \checkmark &Automated  &SegNet& 0.801   \\ 
 Ours  & \checkmark  &    $\times$ &Automated   &SegFormer& 0.813     \\
Ours   & \checkmark  &    \checkmark &Automated   &SegFormer& 0.844   \\

Ours  & \checkmark &  \checkmark &Hand-labeled &SegFormer& 0.878    \\

Ours    & \checkmark &  \checkmark &Automated &U-Net& 0.813 	    \\ 
Ours  &  \checkmark   & \checkmark &Hand-labeled &U-Net& 0.849     \\
Ours    & \checkmark &  \checkmark &Automated &SegNet& 0.824 	    \\ 
Ours  &  \checkmark   & \checkmark &Hand-labeled &SegNet& 0.862    \\

\hline

\end{tabular}
\label{tab:results_mix}
\end{table}

Table~\ref{tab:results_mix} provides a summary of the performance of different methods under various input and model configurations in terms of IoU scores. Overall, the results show that incorporating depth information improves the accuracy of free-space segmentation compared to using only RGB images. We observe that our method, using the aforementioned unsupervised annotation framework, outperforms the corresponding SegNet-based method from~\cite{b2} while also maintaining comparable performance versus its supervised counterparts. Moreover, our approach demonstrates considerable adaptability, by performing well when fine-tuning with different model architectures i.e. U-Net and SegNet.

\subsection{Impact of the number of training data used }

\begin{figure}[h]
    \centering
    \includegraphics[width=10cm]{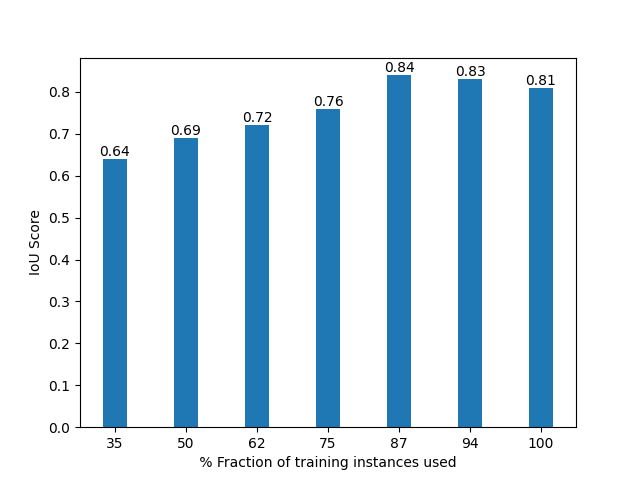}
    \caption{Method's performance for different number of training instances}
    \label{fig:posdata}
\end{figure}

Figure~\ref{fig:posdata} illustrates the correlation between the number of training instances and the accuracy in free-space segmentation. The number of positive data used, plays a crucial role in the learning process of our method since, it helps the algorithm learn a wider range of patterns and variations associated with traversable areas. 
We notice that initially, as the number of instances increases, the IoU scores show a significant improvement. However, beyond the threshold of 1800 images (84\%), it is observed that a further increase of the number of instances has a diminishing impact on the IoU scores. These findings indicate that even with a limited dataset, the proposed method achieves satisfactory performance, reinforced by the substantial ability of the SegFormer to capture complex spatial dependencies.

\subsection{Effect of the number of clusters }

\begin{figure}[h]
    \centering
    \includegraphics[width=10cm]{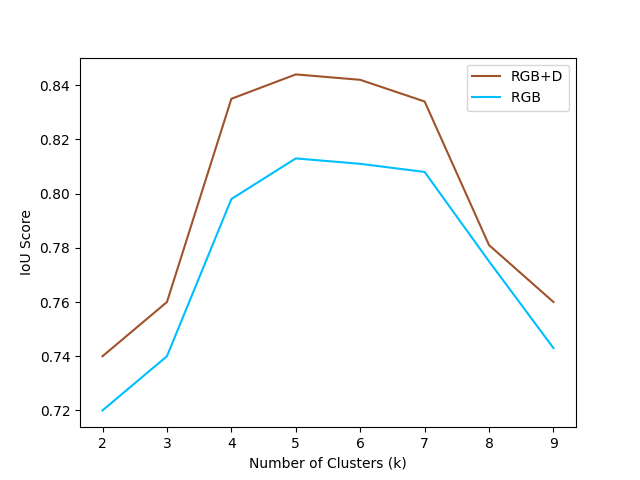}
    \caption{Method's performance for different number of clusters}
    \label{fig:clusters}
\end{figure}

Figure~\ref{fig:clusters} illustrates the relationship between the IoU performance and the number of clusters used in k-means. Increasing the number of clusters leads to improved segmentation performance, ought to the gradually increasing capability of the algorithm to capture depth and texture variations. It should also be remarked that after a certain number of clusters, for instance k = 6, we note a steady degradation in performance due to over-segmentation caused by the fine-grained clusters, capturing noise and redundant details.

\subsection{Qualitative results}
\begin{figure}[h]
    \centering
    \includegraphics[width=15cm]{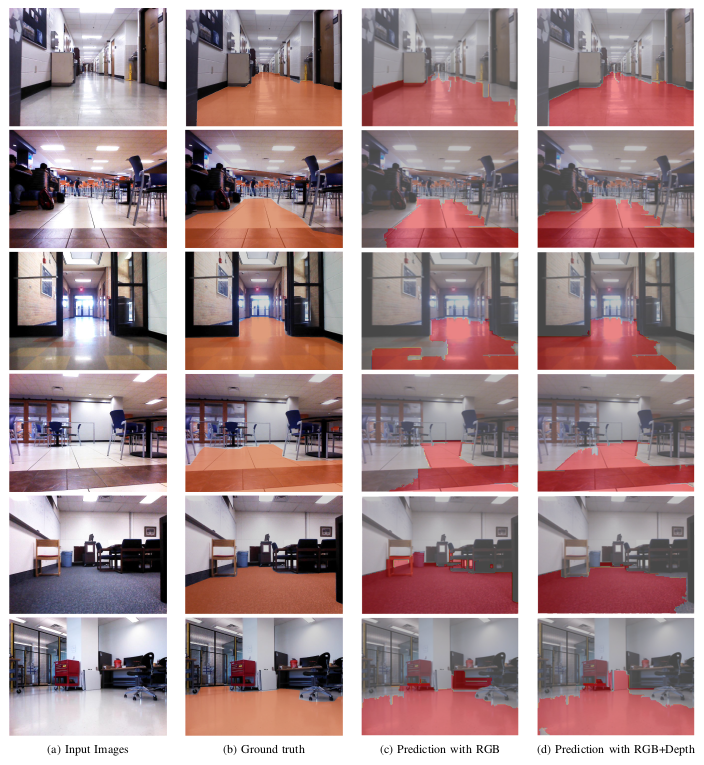}
    \caption{Illustrative examples of the method’s performance, last two rows depict erroneous results}
    \label{fig:qual}
\end{figure}

Figure~\ref{fig:qual} presents some qualitative results on the test dataset. We compare the performance of our method, We compare the performance of our method, that uses both RGB and depth
information during training to generate masks, against the method presented by~\cite{b2} and only uses RGB. Overall, it is noted that incorporating depth information enhances the accuracy of free-space prediction compared to relying solely on RGB data. That is because depth information can provide valuable cues for free-space segmentation in the narrow passages along with some depth discontinuities, that help distinguishing the objects' edges and boundaries from the surrounding free-space regions. 
SegFormer's proficiency to effectively capture the long-range dependencies and spatial information, along with the features of the pretrained DPT within in the input data, is of instrumental importance. It enables the proposed algorithm to accurately differentiate between free-space regions and obstacles, even in challenging scenarios with objects, doors, and varying textures. 

Besides, the use of positive instances during training, allowed the algorithm to prioritize the distinctive features and areas depicting free-space regions. However, some erroneous, false positive predictions were noted (last two rows in Figure~\ref{fig:qual}) as well. This can be attributed to the presence of certain objects, which creates visual similarities with free space regions, consequently leading to misclassifications. 

\begin{figure}
  \centering
  \begin{minipage}[b]{0.43\linewidth}
    \centering
    \includegraphics[width=\linewidth]{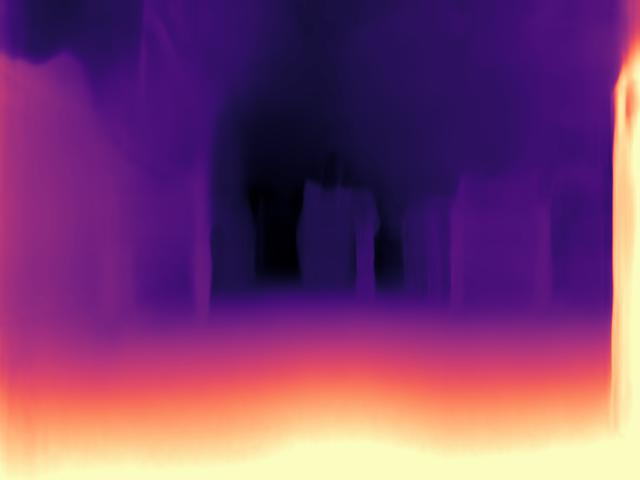}
    %\caption{}
    %\label{fig:image1}
  \end{minipage}
  \hfill
  \begin{minipage}[b]{0.43\linewidth}
    \centering
    \includegraphics[width=\linewidth]{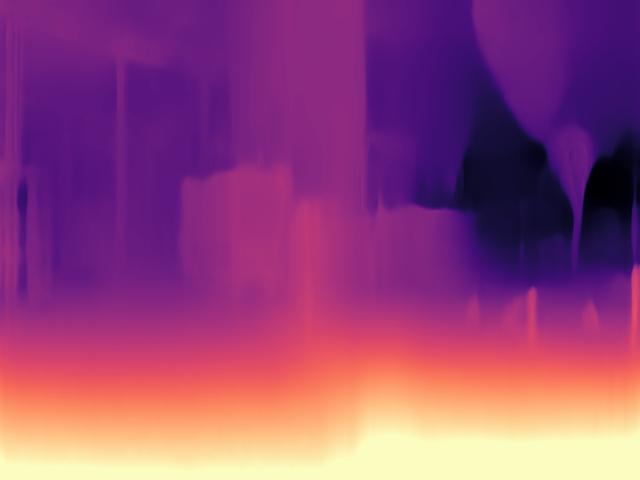}
    %\caption{}
   % \label{fig:image2}
  \end{minipage}
  \caption{Depth maps for the wrong predictions (for rows 5 and 6 respectively) of Figure~\ref{fig:qual}}
  \label{fig:depth_wrong}
\end{figure}

Upon examining the corresponding depth maps in Figure~\ref{fig:depth_wrong}, it becomes evident that the accurate information provided by the depth map does not have an impact on the observed misclassifications. Hence, we can confidently conclude that these errors are a result of our method's performance. More specifically, in the example of row 5/Figure~\ref{fig:qual}, the error can be attributed to the fact that the algorithm does not consider the height of the encountered furniture (in this case the instructor's lectern), which highlights the challenge of ambiguous semantics.
Also, with respect to the sixth row of Figure~\ref{fig:qual}, the tile positioned against the wall leads to a misunderstanding of texture, because it resembles a traversable surface that the algorithm has been previously trained on.

\section{Conclusions and Future Work}

This paper explores the significance of linking navigable regions with areas of higher depth for the purpose of free-space segmentation. We use an efficient automated masking technique, which utilizes textural homogeneity, depth uniformity and positive scenes to generate meaningful segments. Moreover, our method associates extracted features from the Dense Prediction Transformer (DPT) with representative anchor points within superpixels generated by the DASP algorithm. Finally, we fine-tune a SegFormer on our custom-collected dataset, by training on positive and testing on challenging instances while witnessing satisfactory performance.

Erroneous classifications established an avenue for future research. One approach to tackle this error would be leveraging the additional context provided by 3D point clouds by considering both visual cues and the physical positions and heights of objects. Also, experimenting with larger and more diverse datasets of indoor environments could contribute to further refining the algorithm's performance and generalizability.

Our error analysis shows that future research should be directed
towards addressing misclassifications due to misleading texture.
The area under the chair (row 5, Fig.~\ref{fig:qual}) and the tile propped against the wall (row 6, Fig.~\ref{fig:qual}) are characteristic examples:
The algorithm correctly correlates the texture of tiles with
traversable space, but fails to take into account the placement
and position of the tile. This is an instance of the more general
problem of how to combine common-sense (and, generally, structured) 
knowledge with machine learning. Future work can explore a
more explicit combination between textural and geometric
features. Such an approach would allow the system to represent
findings such as 
"tiles propped up are obstacles''
and
"depending on height, overhead obstacles might make the area under them non-traversable"
without jeopardizing the generalization that
"tiles are normally laid over traversable areas''.

\section*{Acknowledgment}

The authors would like to thank Vasileios Kitsakis for his valuable insight and feedback.

%Bibliography
\bibliographystyle{unsrt}  
\bibliography{references}

\end{document}